# USING FRACTAL DIMENSION TO PREDICT THE RISK OF INTRA-CRANIAL ANEURYSM RUPTURE WITH MACHINE LEARNING

**Pradyumna Elavarthi**[1], **Anca Ralescu**[1], **Mark D. Johnson**[2], and **Charles J. Prestigiacomo**[2]

[1]Department of Computer Science, University of Cincinnati
[2]Neurosurgery, College of Medicine, University of Cincinnati

September 27, 2024

## Abstract

Intracranial aneurysms (IAs) that rupture result in significant morbidity and mortality. While traditional risk models such as the PHASES score are useful in clinical decision-making, machine learning (ML) models offer the potential to provide more accuracy. In this study, we compared the performance of four different machine learning algorithms—Random Forest (RF), XGBoost (XGB), Support Vector Machine (SVM), and Multi-Layer Perceptron (MLP)—on clinical and radiographic features to predict rupture status of intracranial aneurysms. Among the models, RF achieved the highest accuracy (85%) with balanced precision and recall, while MLP had the lowest overall performance (accuracy of 63%). Fractal dimension ranked as the most important feature for model performance across all models.

## 1 Introduction

The rupture of intracranial aneurysms (IAs) and subsequent subarachnoid hemorrhage carry high rates of mortality and disability [1]. While only a minority of IAs rupture [2][3], predicting which aneurysms are at the greatest risk is crucial for informing clinical management. The relationship between IA size and rupture risk has long been established [2][4], but more recent studies suggest that size alone is insufficient for determining rupture risk, with small aneurysms (<7 mm) comprising a significant proportion of ruptured IAs [5][6]. In an attempt to gain a more holistic representation of an aneurysm's morphology, shape ratios and morphometric parameters have been investigated with variable results [7][8][9]. The PHASES score, which incorporates both clinical and radiographic factors, is widely used to assess IA rupture risk [3][10]. However, its performance can be limited, with significant weight being given to size and location within the model.

With excitement in the medical community around artificial intelligence, machine learning techniques have been applied by several groups to correlate clinical and radiographic variables with rupture status. Many of these studies focus on regression-based models, which can lack performance in complex nonlinear relationships [11][12][13]. ML models, such as Random Forests (RF), Support Vector Machines (SVM), and neural networks, have shown promise in handling complex nonlinear relationships [11][14]. In this study, we applied various ML techniques, including RF, XGBoost (XGB), SVM, and Multi-Layer Perceptron (MLP), to a dataset of clinical and radiographic variables to assess their ability to predict IAs rupture status. We compared the performance of these models to determine which provides the best predictive power and evaluate how they compare to existing clinical scores like PHASES.

## 2 Methods

### 2.1 Dataset and Feature Extraction

The dataset used in this study consisted of 178 samples with 58 features, excluding the rupture status. These features included clinical parameters such as age, sex, gender, location of the aneurysm, hypertension status, whether the patient has multiple aneurysms, and morphological features corresponding to the geometry of the aneurysm. The morphological features included sphericity, bifurcation or sidewall location, undulation index, sa:vol ratio, lacunarity, size (mm), dome height (mm), maximum Feret diameter, among others. Additionally, a novel feature, fractal dimension (Minkowski Dimension), was incorporated, as it has been shown to correlate with rupture status in a recent study [9].

### 2.2 Modeling

To model the data, we removed outliers that were more than two standard deviations away from the mean and imputed them with the median value of the corresponding feature. We also found that several features were correlated with each other with respect to the target variable, as shown in Figure 1, and dropped features with high correlation (>0.8) to reduce redundancy. Pairwise plots showing the correlations are presented in the 1.

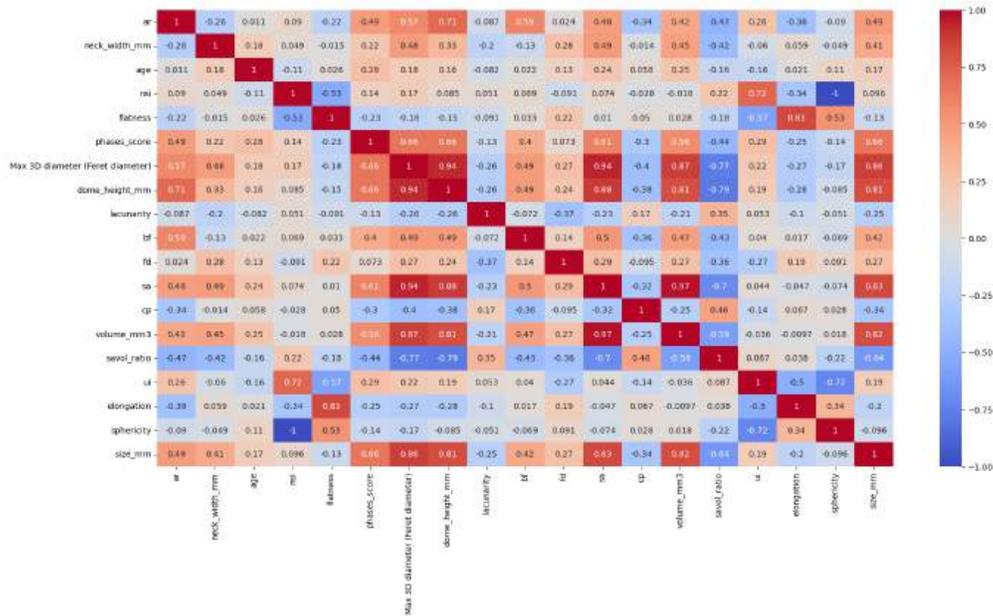

Figure 1: Pairwise correlation plot for the features used in this study

For modeling, we used supervised machine learning models such as Random Forest, XGBoost, SVM, and MLP to predict the rupture status. We applied grid search to fine-tune the hyperparameters for each model and increased the class weights to address the imbalance in the data.

## 3 Statistical Analysis

To minimize the effects of variability inherent in model performance, each model was evaluated across five iterations of 5-fold cross-validation. The final Area Under the Curve (AUC) score for each model was presented as the median of the five AUC scores generated in each iteration. The raw AUC values were recorded in a dedicated column, and the overall performance metrics (accuracy, precision, recall, and F1-score) were extracted using predefined metrics from the Scikit-learn library.

The Wilcoxon signed-rank test was used to compare the AUC scores between model pairs. The Wilcoxon signed-rank test was selected due to its non-parametric nature, ideal for comparing matched-pair data without the assumption of normality.

For model-dependent statistics, the repetition that resulted in the median AUC score was selected to analyze other evaluation metrics, including precision, recall, and F1-score. When multiple repetitions achieved the same median score, the first instance was selected for further analysis.

## 4 Results

The dataset consisted of several clinical and morphological features related to aneurysms, including patient demographics, aneurysm location, and rupture status. The average age of the patients in the dataset was 59.3 years, with aneurysm sizes ranging from small to large, averaging 5.89 mm. The distribution of the target variable, rupture status, was moderately unbalanced, with 63% of aneurysms being classified as unruptured and 37% classified as ruptured. The most common locations were the internal carotid artery (ICA) (n = 42), followed by the anterior communicating artery (AComm) (n = 41), middle cerebral artery (MCA) (n = 32), posterior communicating artery (PComm) (n = 25), basilar artery (n = 17), and posterior inferior cerebellar artery (PICA) (n = 11). Additional locations included the distal anterior cerebral artery (n = 4), anterior inferior cerebellar artery (n = 2), vertebral artery (n = 2), superior cerebellar artery (n = 1), and posterior cerebral artery (n = 1).

### 4.1 Accuracy

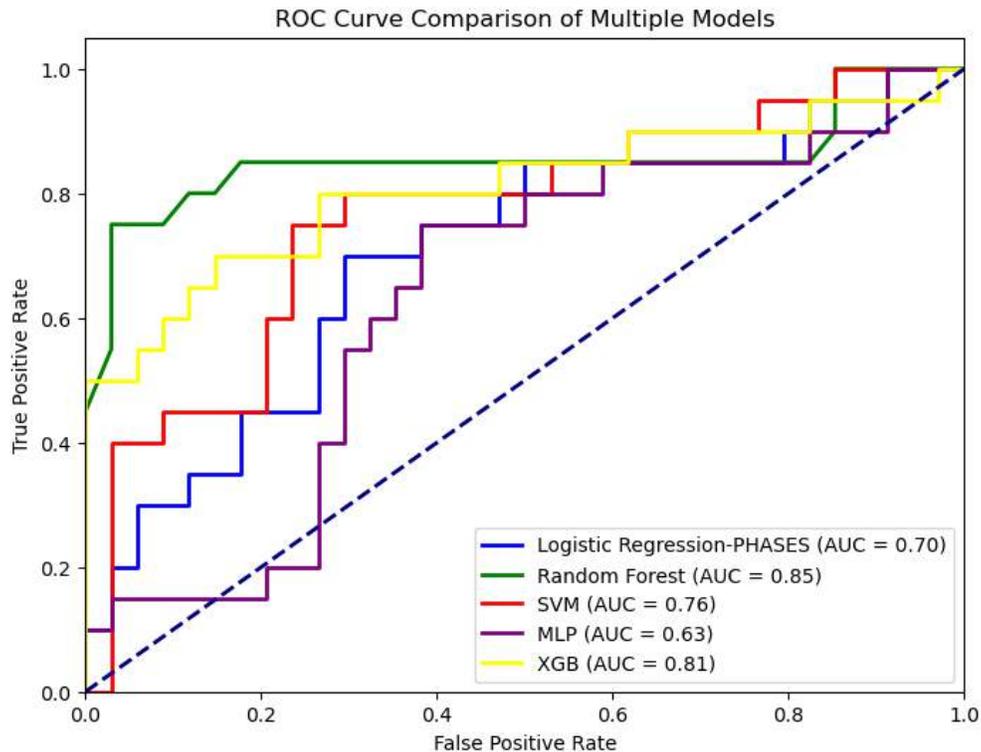

Figure 2: ROC curves for different models used in this study.

As seen in Figure 2, the ROC curve shows that the Random Forest model achieved the highest AUC value of 0.85, while the MLP model had the lowest AUC at 0.63. The accuracy and other performance metrics for each model are summarized in Table 1. XGBoost had the highest recall (0.80), followed closely by Random Forest (0.75), suggesting that both models perform well in detecting true positive ruptures. However, Random Forest demonstrates the highest overall accuracy (0.85), indicating that it performs well in both detecting ruptured aneurysms and correctly identifying non-ruptured cases.

The AUC for the Random Forest model was significantly higher than that obtained through SVM (p-value = 0.031) and MLP (p-value = 0.016). In contrast, the Random Forest model and XGBoost showed no significant difference in performance, (p-value = 0.144) suggesting similar discriminative ability. The comparison between SVM and MLP did not reveal a significant difference in performance, p-value = 0.089. In summary, the Random Forest model outperforms both SVM and MLP, but showed no significant difference in performance compared to XGBoost.

| Model | Accuracy | Precision | Recall | F1-Score |
|---|---|---|---|---|
| Logistic Regression (PHASES) | 0.70 | 0.75 | 0.30 | 0.43 |
| Random Forest | **0.85** | **0.83** | 0.75 | **0.79** |
| SVM | 0.69 | 0.59 | 0.50 | 0.54 |
| MLP | 0.65 | 0.52 | 0.55 | 0.54 |
| XGBoost | 0.76 | 0.64 | **0.80** | 0.71 |

Table 1: Performance metrics for the models used in this study.

### 4.2 Feature Importance

Analysis of individual feature importance across three machine learning models: Random Forest (RF), Support Vector Machine (SVM), and XGBoost, is shown in Figure 3. Each model assigned variable importance to clinical and radiographic features, showing variations in how these models interpret the data.

All three models ranked the *fractal dimension (fd)* as the most important feature. Beyond this homogeneity around fd, the rank and combination of other features varied. XGBoost also identified *aneurysm size*, *total number of intracranial aneurysms (IA)*, and *flatness* as key predictors. Similarly, SVM, highlighted *lacunarity*, *sphericity*, and *total number of IAs*, again emphasizing holistic shape parameters. The Random Forest model further emphasized the importance of shape features including *undulation index (ui)* and *aspect ratio (ar)*.

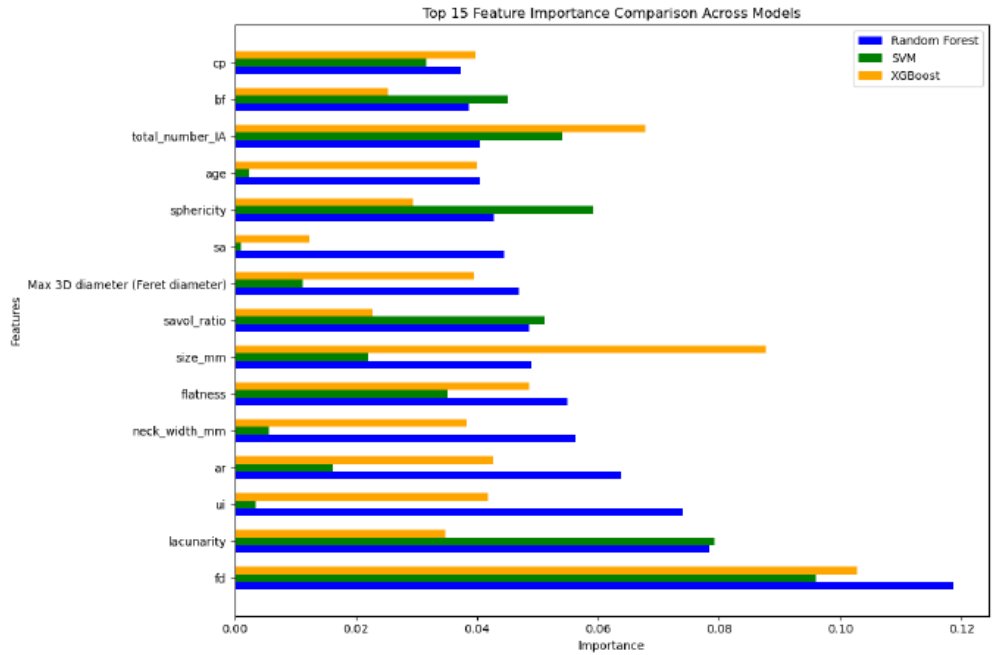

Figure 3: Feature importance rankings for Random Forest, SVM, and XGBoost models.

## 5 Discussion

In this study, we applied three machine learning models—Random Forest (RF), Support Vector Machine (SVM), and XGBoost—to predict aneurysm rupture risk based on clinical and morphological features. All models demonstrated reasonable predictive capabilities, with Random Forest achieving the highest AUC. Consistently across models, the fractal dimension was identified as the most significant predictor, highlighting its role in assessing rupture status.

When compared to models in the literature, our results align with previous studies that have utilized Random Forest and XGBoost for aneurysm rupture status prediction. For example, prior research has demonstrated the strong performance of tree-based models in handling complex, non-linear data [15][16][17]. However, our SVM model, while performing moderately well, did not outperform Random Forest or XGBoost, which is consistent with studies indicating that SVMs may struggle with highly imbalanced datasets or non-linear feature relationships in insufficient medical data.

The prominence of morphological features such as fractal dimension, lacunarity, and sphericity in our models underscores the critical role of aneurysm geometry in classification [7]. It is also important to note that aneurysm size and total number of aneurysms also performed well, particularly in XGBoost. The emphasis on fractal dimension may be due to its ability to capture complex shape irregularities that affect fluid flow, which may provide a more nuanced understanding of rupture risk than size alone. [9]

Despite the promising results, several limitations should be acknowledged. First, the dataset used in this study was relatively small and imbalanced, which could have affected the performance of certain models, particularly SVM. Additionally, while we utilized cross-validation to mitigate overfitting, external validation on a larger and more diverse dataset is necessary to confirm the generalizability of these findings. Nevertheless, this study demonstrates the potential of machine learning models to improve aneurysm classification and guide clinical decision-making.

# 6 Conclusion

In this study, we explored the predictive potential of three machine learning models—Random Forest (RF), Support Vector Machine (SVM), and XGBoost—for assessing the risk of aneurysm rupture. By incorporating both clinical and morphological features, we aimed to identify the most significant predictors of rupture status and compare the performance of these models. Among the models, Random Forest emerged as the top performer, consistently identifying fractal dimension, lacunarity, and other geometric features as key factors in aneurysm rupture risk. Our findings highlight the importance of incorporating advanced geometric descriptors such as fractal dimension and sphericity, which may offer more nuanced insights compared to traditional clinical metrics like aneurysm size alone.

# 7 Appendix

In this appendix, we provide a detailed list of the features used in our research study on intracranial aneurysm rupture and their corresponding importances for each model. Table ?? includes both the abbreviated feature names and their corresponding full descriptions, which offer insights into the morphological, clinical, and demographic factors considered in the analysis.

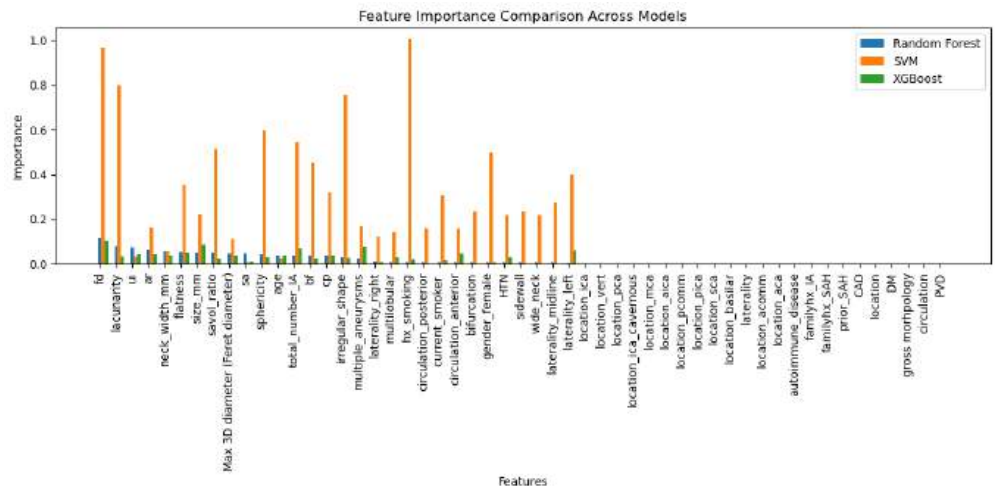

Figure 4: Feature importance of all features used in modelling.

Table 2: Features and their full names used in the research study on intracranial aneurysm rupture.

| Index | Feature | Description |
| --- | --- | --- |
| 1 | age | Age |
| 2 | ar | Aspect Ratio |
| 3 | autoimmune_disease | Autoimmune Disease |
| 4 | bf | Bulge Factor |
| 5 | bifurcation | Bifurcation |
| 6 | CAD | Coronary Artery Disease |
| 7 | circulation | Circulation |
| 8 | circulation_anterior | Circulation (Anterior) |
| 9 | circulation_posterior | Circulation (Posterior) |
| 10 | cp | Compactness |
| 11 | current_smoker | Current Smoker |
| 12 | DM | Diabetes Mellitus |
| 13 | familyhx_IA | Family History of Intracranial Aneurysm |
| 14 | familyhx_SAH | Family History of Subarachnoid Hemorrhage |
| 15 | fd | Fractal Dimension |
| 16 | flatness | Flatness |
| 17 | gender_female | Gender (Female:1 Male:0) |
| 18 | gross_morphology | Gross Morphology |
| 19 | HTN | Hypertension |
| 20 | hx_smoking | History of Smoking |
| 21 | irregular_shape | Irregular Shape |
| 22 | lacunarity | Lacunarity |
| 23 | laterality | Laterality |
| 24 | laterality_left | Laterality (Left) |
| 25 | laterality_midline | Laterality (Midline) |
| 26 | laterality_right | Laterality (Right) |
| 27 | location | Location |
| 28 | location_aca | Location (Anterior Cerebral Artery) |
| 29 | location_acomm | Location (Anterior Communicating Artery) |
| 30 | location_aica | Location (Anterior Inferior Cerebellar Artery) |
| 31 | location_basilar | Location (Basilar Artery) |
| 32 | location_ica | Location (Internal Carotid Artery) |
| 33 | location_ica_cavernous | Location (ICA Cavernous) |
| 34 | location_mca | Location (Middle Cerebral Artery) |
| 35 | location_pca | Location (Posterior Cerebral Artery) |
| 36 | location_pcomm | Location (Posterior Communicating Artery) |
| 37 | location_pica | Location (Posterior Inferior Cerebellar Artery) |
| 38 | location_sca | Location (Superior Cerebellar Artery) |
| 39 | location_vert | Location (Vertebral Artery) |
| 40 | Max 3D diameter (Feret diameter) | Maximum 3D Diameter (Feret Diameter) |
| 41 | multilobular | Multilobular Morphology |
| 42 | multiple_aneurysms | Multiple Aneurysms |
| 43 | neck_width_mm | Neck Width (mm) |
| 44 | prior_SAH | Prior Subarachnoid Hemorrhage |
| 45 | PVD | Peripheral Vascular Disease |
| 46 | sa | Surface Area |
| 47 | savol_ratio | Surface Area to Volume Ratio |
| 48 | sidewall | Sidewall Aneurysm |
| 49 | size_mm | Aneurysm Size (mm) |
| 50 | sphericity | Sphericity |
| 51 | total_number_IA | Total Number of Intracranial Aneurysms |
| 52 | ui | Undulation Index |
| 53 | wide_neck | Wide Neck |